\documentclass{article} 

\usepackage{iclr2027_arxiv,times}

\usepackage[utf8]{inputenc}
\usepackage[stable]{footmisc}
\usepackage{amsmath}  
\usepackage{graphicx} 
\usepackage{amsthm}
\usepackage{tikz-cd} 
\usepackage{amssymb}
\usepackage{bm}
\usepackage{bbm}
\usepackage{booktabs}
\usepackage{pifont}
\usepackage{xcolor}
\usepackage{array}
\usepackage{makecell}
\usepackage{dsfont}
\usepackage{graphicx}
\usepackage{subcaption}
\usepackage{blindtext}
\usepackage{url}
\usepackage{algorithm}
\usepackage{algorithmic}
\usepackage{thmtools}
\usepackage{thm-restate}
\usepackage{setspace}
\usepackage{enumitem}
\usepackage{mathtools}
\usepackage{xcolor}
\usepackage[final]{hyperref} 
\hypersetup{
	colorlinks=true,       
	linkcolor=blue,        
	citecolor=blue,        
	filecolor=magenta,     
	urlcolor=blue         
}
\usepackage{cleveref}
\usepackage{tcolorbox}
\usepackage{tabularx}

\newtheorem{theorem}{Theorem}[section]

\theoremstyle{definition}

\theoremstyle{definition}

\newtheorem{remark}[theorem]{Remark}

\numberwithin{equation}{section}

\newcommand{\R}{\mathbb{R}}

\renewcommand{\S}{\mathbb{S}}

\renewcommand{\P}{\mathsf{P}}

\makeatletter
\def\url@leostyle{%
	\@ifundefined{selectfont}{\def\UrlFont{\sf}}{\def\UrlFont{\small\ttfamily}}}
\makeatother
\definecolor{darkgreen}{rgb}{0,0.4,0}

\title{The Role of Feed-Forward Layers in Transformer Dynamics}

\author{Thomas Jacob Maranzatto, Semih Akkoc \& Sennur Ulukus \\
Department of Electrical and Computer Engineering \\
University of Maryland \\
College Park, MD 20742, USA \\
\texttt{\{tmaran,akkoc,ulukus\}@umd.edu}
}

\iclrfinalcopy 

\begin{document}
\maketitle
\begin{abstract}
  We study the dynamical behavior of tokens in transformers from a control-theoretic perspective.  Our model includes the feed-forward layer present after the self-attention mechanism, with the self-attention mechanism interpreted as an interacting particle system and the feed-forward layer as an independent control.  Our main theoretical result establishes that the feed-forward network can steer the tokens arbitrarily close to consensus regardless of the key, query, and value matrices.  Our result are easily extended to convergence to many clusters and to multi-head attention.  We conduct numerical experiments to verify our results, and compare thresholding behavior from our theory to real-world LLMs.
\end{abstract}

\section{Introduction}
Since their introduction less than a decade ago~\citep{vaswani2017attention}, transformers have become a foundational architecture for machine learning and AI~\citep{bert,dosovitskiy2021an,gemini2023,gpt,llama, kovalevskiy2024alphafold}.  Despite the rapid progress seen in deployed systems, the theory of transformers is relatively nascent.  There has been a recent push to analyze transformer inference-time behavior as a continuous time coupled ODE system~\citep{sander2022sinkformers,geshkovski2023mathematical,bruno2024emergence,castin2025unified,burger2025analysis,bruno2025multiscale,alcalde2026quantifying, li2026diverse}, which has attracted theorists and practitioners for its explainability and rich behavior.   We also note the recent work bringing universal approximation theory to transformer analysis~\citep{furuya2024transformersuniversalincontextlearners, li2025transformersmeetincontextlearning, gumaan2025universalapproximationtheoremsinglelayer}.

This paper takes the coupled ODE approach to analyzing the inference time behavior of transformers.  The primary model we consider is inspired by~\citep[Equation 2.8]{geshkovski2023mathematical}, and interprets the transformer as an interacting particle system evolving according to the following ODE,
\begin{equation}
    \label{FSA}\tag{FSA}
    \dot x_i(t) = \underbrace{\P_{x_i(t)}^\perp \left( \frac{1}{Z_i(t)}\sum_{j=1}^n e^{\beta \langle x_i(t), Ax_j(t)\rangle} V x_j(t)\right)}_{\text{self-attention}} + \underbrace{\P_{x_i(t)}^\perp \left( \sum_{k=1}^\ell w_k(t) \sigma(a_k(t) x_i(t) + b_k(t)) \right)}_{\text{feed-forward layer}}
\end{equation}
Our \textbf{feed-forward self-attention}~\eqref{FSA} model has a number of components.  $x_i(t) \in \S^{d-1}$ is interpreted as the output of the $t$'th transformer layer for the $i$'th token.  $\P_x^\perp = I - xx^\top$ is the projection matrix that sends vector $y \in \R^d$ to $y' \in T_x \S^{d-1}$. The matrices $A,V \in \R^{d \times d}$ are parameter matrices, with $V$ the value matrix and $A = Q^\top K$ the combined query and key matrices.  $\beta$ is an inverse temperature that controls the token velocities and $Z_i(t) = \sum_ke^{\beta \langle x_i(t), Ax_k(t)\rangle}$ is a normalizing factor, sometimes called a partition function.  For the feed-forward layer, $\ell$ denotes the number of hidden neurons, $w_k(t), a_k(t) \in \R^{d \times d}$ are neuron and layer dependent weights, $b_k(t) \in  \R^{d}$ is a layer-dependent bias, and $\sigma$ is a function applied component-wise to the input vector.  Throughout, $\|v\|$ denotes the Euclidean norm of a vector, $\|M\|_{op}$ denotes the operator norm of a matrix, and we will use the following shorthand for the projected feed-forward layer acting on an input $x$: \[u_x(t) := \P_{x(t)}^\perp \left( \sum_{k=1}^\ell w_k(t) \sigma(a_k(t) x(t) + b_k(t)) \right).\]

We are aware of two other works which incorporate the feed-forward layer into the interacting particle system perspective~\citep{alvarez2026perceptrons, Adu2024feedforward}. Both works consider ReLU/GeLU activations, and perform their analysis in the mean-field PDE limit.  The works differ in their goals.  In~\citep{alvarez2026perceptrons}, the authors characterize the possible stationary points of the interacting particle system, and view the feed-forward network as a parametrized external field.  In~\citep{Adu2024feedforward} the authors show the continuum limit of~\eqref{FSA} can approximate any sequence of vector fields and probability measures on $\S^{d-1}$.  To prove their approximation result, the authors formulate a control problem for all parameters in both the self-attention block and feed-forward layer.  Our work differs from these previous works in a few ways.  First, \textit{our goal is to quantify how the introduction of the feed-forward layer impacts the long-term dynamics of~\eqref{FSA}.}  We also want to quantify how strong the external field must be to force the system to consensus.  As far as we are aware, the works above do not quantify the relative impact of the feed-forward layer compared to the self-attention layer.  Our approach is similar to~\citep{Adu2024feedforward} in that we view the system as a  control problem, though our methodology is adapted from the opinion dynamics literature~\citep{Caponigro2015}.  We design $\sigma, w,a,b$ in a natural way that guarantees $x_i(t) \in  B_\epsilon(x_0)$ for almost all $i$ and for all $A,V$, under some suitable conditions on the parameters above.  Here $B_\epsilon(x_0) = \{x \in \S^{d-1}: \frac{1}{2}\|x - x_0\|^2 \leq \epsilon\}$.  Our main result is presented in the theorem below, where $\|A\|_{\operatorname{op}}$ denotes the standard operator norm.

\begin{theorem}\label{thm:main_result}
    For any $\epsilon > 0$, $A,V$ and $x_0 \in \mathbb{S}^{d-1}$ there exist a choice of $\sigma, w(t), a(t),$ and $b(t)$ such that for almost all initial conditions $\{x_i(0)\}_1^n \in (\mathbb{S}^{d-1})^n$,   the system~\eqref{FSA} with a one-neuron feed-forward layer satisfies,
    \[x_i(t) \in B_\epsilon(x_0)\cap \mathbb{S}^{d-1} \quad \forall i, \forall t \geq t_0.\]
    for some $t_0 < \infty$.  Furthermore, the choice of parameters above satisfy $\inf\|u_x(t)\|  >  \|V\|_{\operatorname{op}}$ for $x \not\in B_\epsilon(x_0) \cup \{-x_0\}$ and $u_x(t)$ is continuous on $\mathbb{S}^{d-1} \setminus \{-x_0\}$. Conversely there exist matrices $A$,$V$ such that no feed-forward layer with $\sup \|u_x(t)\| < \|V\|_{\operatorname{op}}$ satisfies $\lim_{t \to\infty} x_i(t) \to x_0$ $\forall i$.
\end{theorem}

Notice that the feed-forward term in~\eqref{FSA} induces a (possibly time-varying) vector field on $\S^{d-1}$ which is independent of the self-attention term.  Theorem~\ref{thm:main_result} demonstrates that the long-time behavior of the system can be steered by this independent vector field, even with only one hidden neuron in the feed-forward layer.  Note if the vector field induced by the feed-forward layer has constant magnitude outside $B_\epsilon(x_0) \cup \{-x_0\}$, then the theorem implies $\|V\|_{\operatorname{op}}$ is a threshold for the feed-forward layer to dominate the dynamics; below this threshold there exist $A$ and $V$ where no feed-forward layer can force synchronization, and above this threshold there is a feed-forward layer that forces tokens to an arbitrarily small neighborhood around $x_0$ for all $A$ and $V$.  In section~\ref{subsec:real_llms} we observe that deployed transformers have feed-forward layers where regions of the induced vector fields exceed the threshold $\|V\|_{\operatorname{op}}$, and later layers which are below this threshold.  This suggests that the `true' dynamics in real systems involve a much richer interplay between the self-attention and feed-forward layers than previous studies have anticipated.  This paper is a first attempt at quantifying and explaining this interplay.

Theorem~\ref{thm:main_result} is stated for long-time convergence to a small ball around  $x_0 \in \S^{d-1}$ for single-head attention.  In remarks~\ref{rem:multihead} and~\ref{rem:multicluster} we outline how our proof can be generalized to multi-cluster convergence for multi-head attention.  Proving convergence to a single cluster is natural given the previous work in similar models~\citep{geshkovski2024dynamic, geshkovski2023mathematical, burger2025analysis, bruno2024emergence}, but is actually undesirable in real transformer architectures.  This is because convergence of all tokens to $x_0$ will always induce the same next-token distribution, effectively turning the transformer into a random token generator. By connecting the feed-forward layer to control theory, we present a generalizable theory that allows analysis beyond token clustering.

\section{A Control-Theoretic Perspective on the Feed-Forward Layer}
\label{sec:control}
Before we prove Theorem~\ref{thm:main_result}, we want to briefly discuss one possible interpretation of the system~\eqref{FSA}.  This interpretation is inspired by opinion dynamics on the sphere~\citep{Caponigro2015}, and is used throughout the proof of our main theorem.  We also use this perspective to extend the analysis to multi-cluster states in the long-time limit.  We begin with the well-known result that feed-forward neural networks are universal function approximators~\citep{hornik1989approximation}.  Since the feed-forward term $u_x(t)$
is a mapping from $\mathbb{S}^{d-1}$ to $\mathbb{T}\mathbb{S}^{d-1}$, we may approximate any function in this family using a feed-forward network.  We interpret the feed-forward layer as control which is learned during training, but fixed during the encoding procedure\footnote{Recall that time $t$ corresponds to the $t$'th layer in the transformer, so the control is allowed to vary in time.  We assume $w(t), a(t), b(t)$ are explicitly known beforehand, though this is a detail that is irrelevant to our analysis.}.  We note that the feed-forward parameters cannot depend on the token positions $\{x_i\}_1^n$, and instead induce a universal external field acting on all tokens.  Indeed, if the feed-forward parameters could depend on the token positions then this layer could simply cancel the self-attention term.

The interpretation of~\eqref{FSA} as a control problem allows us to view $u_i(t)$ as the maximizer of an optimization problem.  Since $u_i(t)$ can approximate any vector field on $\S^{d-1}$, we formally extend $u_i(t)$ to actually attain any vector field on the sphere.  As we see below, in some cases this formalization is unnecessary and we may be able to choose $\sigma, w, a, b$ explicitly to satisfy the desired control problem.  Given any desired long-time behavior we can design a time-dependent optimization problem of the form,
\begin{equation} \label{eq:gen_opt}
    u_x(t) =  \{\max f_t(v(t)) \quad \text{ for } v(t) \in \mathbb{R}^d \text{ such that } \langle v(t), x \rangle = 0, \text{ and } \|v(t)\| \leq  k \},\end{equation}
where we have included the constraint $\|v(t)\| \leq  k$ to ensure that all vectors are bounded. For our problem we set the number of hidden units $\ell = 1$.  Let $e_1 = (1,0,\ldots,0)$ and choose the fixed parameters $a(t) = e_1 x_0^\top$, $b(t) \equiv \mathbf 0$, $w(t) = kx_0 e_1^\top$ with $k > \|V\|_{op}$, and $$\sigma(s) = \begin{cases}
    \frac{1}{\sqrt{1 - s^2}},& \quad s < 1- \epsilon\\
    \frac{1-s}{\epsilon\sqrt{ 2\epsilon - \epsilon^2}}  &\quad 1- \epsilon \leq s \leq 1.
\end{cases}$$   If $s = -1$ then set $\sigma(-1) = 0$.  Notice if $x \notin B_{\epsilon}(x_0) \cup \{-x_0\}$ then the vector $u_x(t)$ is constant ($u_x(t) = u_x$) and is the solution to the problem
    \begin{equation} \label{eq:optproblem}\tag{Problem 1}
    \max \langle v, x_0 \rangle \quad \text{ for } v \in \mathbb{R}^d \text{ such that } \langle v, x \rangle = 0, \text{ and } \|v\| \leq  k.\end{equation}
    Indeed, off of the set $B_{\epsilon}(x_0) \cup \{-x_0\}$ we have,
    \begin{align}
        u_x(t) = \P_{x}\big(kx_0 e_1^\top \sigma((\langle x, x_0\rangle), 0,\ldots 0)^\top)\big) = \frac{k\P_{x} (x_0)}{\|\P_{x} (x_0)\|}.
    \end{align}
    In the case that $x = -x_0$, then $u_x(t) = 0$.  Furthermore, $u_x$ is continuous on $\mathbb{S}^{d-1} \setminus \{-x_0\}$ and $u_{x_0} = 0$.  The activation is regularized near $s=1$ by interpolating from $\frac{1}{\sqrt{2\epsilon - \epsilon^2}}$ to $0$ which preserves continuity near the target $x_0$.  We could have removed the $s \geq 1- \epsilon$ case and handled the discontinuity at $x_0$ by writing a corresponding Filippov differential equation, but we believe the current presentation is cleaner.

\section{Proof of Theorem~\ref{thm:main_result}.}

\textbf{Existence of a synchronizing control.}
    
    Choose $a, b, \sigma, w$ as in the discussion above -- these parameters satisfy Equation~\eqref{eq:optproblem} for $x \not \in B_\epsilon(x_0) \cup \{-x_0\}$.  Set $k > \|V\|_{\operatorname{op}}$, and for now assume $x_i(0) \not\in B_{\epsilon}(x_0) \cup \{- x_0\}$.  Using the shorthand $u_i(t) := u_{x_i}(t)$ we have,
    \begin{align}
        \langle u_i, x_0 \rangle &= \langle u_i - \langle u_i, x_i \rangle x_i, x_0\rangle\\
        &= \langle u_i, x_0 \rangle - \langle u_i, x_i \rangle\langle x_i, x_0\rangle\\
        &= \langle u_i, x_0 - \langle x_i, x_0\rangle x_i  \rangle \\
        &= \|u_i\| \|\P_{x_i}(x_0) \|
    \end{align}

    Where the equality is realized in Cauchy-Schwarz since $u_i$ is the maximizer of~\eqref{eq:optproblem}.  Define $\alpha_i(t) = \frac{1}{Z_i(t)} \sum_j e^{\beta \langle x_i(t), Ax_j(t)\rangle}Vx_j(t)$.  We can now compute,
    \begin{align}
        \frac{d}{dt}(1 - \langle x_i, x_0 \rangle )&= - \langle \dot x_i, x_0 \rangle \\
        &= -\langle \alpha_i, x_0\rangle + \langle \alpha_i, x_i \rangle\langle x_i, x_0 \rangle - \langle u_i, x_0 \rangle \\
        &=-\langle \alpha_i, x_0\rangle + \langle \alpha_i, x_i \rangle\langle x_i, x_0 \rangle -\|u_i\| \|\P_{x_i}(x_0) \| \\
        &= -\langle \alpha_i, x_0 - \langle x_i, x_0 \rangle x_i \rangle -\|u_i\| \|\P_{x_i}(x_0) \| \\
        &\leq \big(\|\alpha_i\| - \|u_i\| \big) \|\P_{x_i}(x_0) \| \\ 
        &= \big(\|\alpha_i\| - \|u_i\| \big) \sqrt{1 - \langle x_i, x_0 \rangle ^2}\\
        &= \big(\|\alpha_i\| - \|u_i\| \big) \sqrt{1 - \langle x_i, x_0 \rangle}\sqrt{1 + \langle x_i, x_0 \rangle}
    \end{align}
    We can bound $\| \alpha_i\|$ in terms of the matrix $V$.  Indeed,
    \begin{equation}
        \|\alpha_i\| = \bigg\| \frac{1}{Z_i} \sum_j e^{\beta \langle x_i, Ax_j\rangle}Vx_j \bigg\| \leq \left|\frac{1}{Z_i} \right| \left|\sum_j e^{\beta  \langle x_i, Ax_j\rangle }\right| \sup_{x\in \S^{d-1}}\|Vx \| \leq \|V\|_{\operatorname{op}}
    \end{equation}
    Let $r_i:= 1 - \langle x_i, x_0\rangle$.  Since we assume $k > k^* := \|V\|_{\operatorname{op}}$ then $r_i$ is non-increasing, and in particular $\langle x_i(t), x_0\rangle \geq \langle x_i(0), x_0\rangle$.  Therefore, if we let $m_i = (k - k^*)\sqrt{1 + \langle x_i(0), x_0\rangle}$ we have

    \begin{equation} \label{eq:diff_ineq}
        \dot r_i(t)\leq- m_i \sqrt {r_i(t)}, \quad x_i(t) \not \in B_\epsilon(x_0)
    \end{equation}
    So that,
    \begin{equation}
        r_i(t) \leq \left(\sqrt{r_i(0)} - \frac{m_it}{2} \right)^2, \quad x_i(t) \not \in B_\epsilon(x_0)
    \end{equation}
    Which implies $x_i(t_0) \in  B_\epsilon (x_0)$ at some time $t_0 < \infty$.  Furthermore Equation~\eqref{eq:diff_ineq} holds on $\partial B_\epsilon(x_0)$, so $B_\epsilon(x_0)$ is forward-invariant and hence $x_i(t) \in  B_\epsilon (x_0)$ for all future times $t > t_0$.  This observation also completes the case $x_i(0) \in B_{\epsilon}(x_0)$.  The case $x_i(t) = -x_0$ is the only setting where $x_i \not\to x_0$ in general, since $-x_0$ may be a fixed point of the dynamics $\dot x_i(t) = \P_{x_i}\alpha_i(t)$.  However if $x_i(0) \not= -x_0$, then for no later time do we have $x_i(t) = -x_0$, since  $\langle x_i(t), x_0 \rangle$ is non-decreasing.
    
\textbf{Non-synchrony for small induced vector fields.}

Above we have given a feed-forward layer with $\|u_i\| >  \|V\|_{\operatorname{op}}$ that is  sufficient to force all $x_i(t_0) \in B_\epsilon(x_0)$ for almost any initial condition.  We now demonstrate that the lower bound on $\|u_i\|$ is in some sense a necessary condition for \textit{any} feed-forward network to force synchrony by considering the setting $V = J :=\begin{psmallmatrix}
    0,& I_{d/2}\\
    -I_{d/2},& 0
\end{psmallmatrix}$ and allowing arbitrary matrix $A$.  Note that for all unit vectors $x$, $\|Vx\| = 1$ and hence $\|V\|_{\operatorname{op}} = 1$

Suppose the feed-forward layers satisfy $k := \sup_{x,t} \|u_x(t)\| < 1$. Assume for sake of contradiction we have $x_i(t) \to x_0$ for all $i$.

Set $\eta = J x_0$. By our assumption it holds that $\alpha_i(t) \to \eta$, and additionally
\[ \| \eta \| = 1, \quad \langle \eta, x_0 \rangle = 0.\]
Define $s_i(t) = \langle x_i(t), \eta \rangle$. Then since $\P_{x_i}$ is symmetric,
\[\dot s_i = \langle \P_{x_i} \alpha_i + u_i, \eta \rangle = \langle \alpha_i, \P_{x_i} \eta \rangle + \langle u_i, \eta \rangle.\]
Since we assume $x_j \to x_0$,
\[\P_{x_i} \eta \to \P_{x_0} \eta = \eta,\]
so that
\[\langle \alpha_i, \P_{x_i} \eta \rangle \to 1.\]
Using our assumption on $u_i$, we can bound the other term as,
\[\langle u_i, \eta \rangle \geq -\|u_i\|\|\eta\| = -\|u_i\| \geq -k.\]
Hence we have
\[\liminf_{t \to \infty} \dot s_i(t) \geq 1-k > 0.\]
In particular for large $t$, $\dot s_i(t) \geq \frac{1-k}{2}$, which contradicts $s_i(t) \to \langle x_0, Jx_0 \rangle = 0.$ \hfill \qed
\begin{remark} \label{rem:almost_any}
    We have stated Theorem~\ref{thm:main_result} for almost any initial configuration of particle configurations.  The obstruction to proving full convergence is the antipodal point $-x_0$, where $\P_{-x_0}(u_i) = \bf0$.  This obstruction is easily overcome if we instead allow multiple neurons in the feed-forward layer of~\eqref{FSA}, as we could construct the field at $-x_0$ to be any non-zero vector.
\end{remark}
\begin{remark}\label{rem:multihead}
    For convenience we have conducted our analysis for single-head attention model~\eqref{FSA}. Our analysis can be extended to the multi-head, time-varying case,
    \begin{equation}
    \label{multi-SA}
    \dot x_i(t) = \P_{x_i(t)}^\perp \left( \sum_{h=1}^H\sum_{j=1}^n \frac{e^{\beta \langle x_i(t), A^h(t)x_j(t)\rangle} V^h(t) x_j(t)}{Z_{i,h}} + \sum_{k=1}^\ell w_k(t) \sigma\left(a_k(t)x_i(t) + b_k(t) \right)\right)
\end{equation}
Letting $\alpha_i(t) := \sum_h \sum_j \frac{e^{\beta \langle x_i(t), A^h(t)x_j(t)\rangle} V^h(t) x_j(t)}{Z_{i,h}}$ and defining $k^* := \sum_h \sup_t \|V^h(t)\|_{\operatorname{op}}$, then the above analysis is readily extended to this general setting, assuming the operator norms are bounded in time.  We will use this model in our analysis of real-world systems below.
\end{remark}
\begin{remark}\label{rem:multicluster}
    We can drive the system to a multiple consensus points $\{x_{\textbf{cons}}^j\}_{j=1}^s$ by modifying~\eqref{eq:optproblem}.  Defining the ball $B_j := \{x \in \mathbb{S}^{d-1} : d_{\mathbb{S}^{d-1}}(x, x_{\textbf{cons}}^j) \leq \frac{1}{2}\min\limits_{k \not = j} d_{\mathbb{S}^{d-1}}(x_{\textbf{cons}}^j, x_{\textbf{cons}}^k)\}$, then the control
    \begin{equation}
    \label{eq:multiple_control}\tag{Problem 2}
        u_x(t) = \begin{cases}\max \langle v, x_j\rangle \quad \text{s.t. } v\perp x, \, \|v\| = k \qquad \text{if } x \in B_j \text{ for some } j
            \\
        \max \langle v, x_0\rangle \quad \text{s.t. } v\perp x, \, \|v\| = k \qquad \text{else}
        \end{cases}
    \end{equation}
ensures $x_i(0) \to \{x_{\textbf{cons}}^h\}$ for some $h$.  Indeed, if $x_i \in B_h$ for some $h$, then the above analysis shows $x_i \to x_{\textbf{cons}}^j$ since $x_i(t) \in B_h$ for all $t$.  Otherwise, the second case in the control problem enforces $x_i \to \partial B_k$ for some $k$ in finite time, at which point $x_i \to x_{\textbf{cons}}^k$.

Some care needs to be taken since the field is discontinuous at each $\pm x_{\textbf{cons}}^j$ and in general $B_i \cap B_j \not = \emptyset$, but we may choose appropriate length $k$ vectors at each point in these sets.  We omit the details, but the proof with an appropriately modified~\eqref{eq:multiple_control} follows the same logic as Theorem~\ref{thm:main_result}.
\end{remark}

\section{Experiments and Empirical Findings}

Theorem~\ref{thm:main_result} gives conditions on the feed-forward layer to induce clustering around $x_0$.  Namely the vector field induced by the feed-forward layer has a minimum magnitude $k^* := \|V\|_{\operatorname{op}}$ required to force convergence to $B_\epsilon(x_0)$.  Our construction provides a field with $\|u_x(t)\| = k > k^*$ for all $x \not\in B_\epsilon(x_0) \cup \{-x_0\}$.  Our counterexample with $V=J$ shows any field with $\|u_x(t)\| < k^*$ for all $x$ will fail to synchronize. Note that these are not necessary and sufficient conditions, as most vector fields satisfy $\inf \|u_x\| < k^* < \sup \|u_x\|$.

This motivates the experiments below.  First we numerically verify Theorem~\ref{thm:main_result} in Section~\ref{subsec:numeric_conf} by varying the strength of $k$ in~\eqref{eq:optproblem} in two settings of $A$ and $V$.  Second we investigate real-world transformers in Section~\ref{subsec:real_llms} by characterizing their layer-by-layer values of $k^*$, studying the vector field induced by the feed-forward layer, and identifying the distribution of $\{\|u_x\|\}_{x \in \S^{d-1}}$.

\subsection{Numerical Confirmation of the Theoretical Results}
\label{subsec:numeric_conf}
In both experiments below, we numerically integrate~\eqref{FSA} with the feed-forward layer induced by~\eqref{eq:optproblem} and send the field to 0 continuously in $B_\epsilon(x_0)$ for very small $\epsilon$.  In all experiments, we initialize $\{x_i(0)\}_1^n$ i.i.d. uniformly on $\S^{d-1}$.  Unless otherwise stated, experiments are performed with $n=100$, $d=10$, and $\beta = 1$.  All experiments in this section use fourth-order Runge--Kutta.

For each trial, we compute $k^{*}=\|V\|_{\mathrm{op}}$ and in the plots we vary $k/k^{*}$. This gives the same normalized horizontal axis across trials even when $V$ changes. Each grid point is averaged over independent initial tokens and targets $x_0$. In the second experiment, $A$ and $V$ are also resampled for each trial.

In the experiments below we plot the two following quantities. The first is the mean pairwise inner product,
\begin{equation}
    R_1(t) := \binom{n}{2}^{-1}\sum_{i<j}\langle x_i(t),x_j(t)\rangle \in [0,1]
\end{equation}
When $R_1=1$ all tokens satisfy $x_i = \mathbf{x}$ for some $\mathbf{x}$, and when $R_1 \approx 0$ the tokens are approximately uniformly distributed. The second quantity is the mean distance to the target $x_0$,
\begin{equation}
    Q_1(t) :=\frac{1}{n}\sum_{i=1}^{n}\|x_i(t)-x_0(t)\|_2 \in [0,2]
\end{equation}

Naturally, $Q_1 = 0$ if $x_i = x_0$ for all $i$, and $Q_1 = 2$ if $x_i = -x_0$.  Note that since $x_i(0)$ is drawn uniformly in our experiments, we expect $Q_1(0) \approx \sqrt{2}$.

We plot both $R_1$ and $Q_1$ at time $t=t_{\mathrm{final}}$ since it is possible for the system to synchronize to some state $x_1\neq x_0$. In addition to these final values, we also report tail-averaged quantities. For each trajectory, the tail average is the arithmetic mean over the final $10\%$ of the recorded numerical time steps, and the tail distance is the average of $Q_1(t)$ over this final portion of the trajectory. We then average this quantity over the independent trials at each value of $k/k^*$. Comparing the final and tail-averaged values provides a check that the dynamics have effectively settled by $t=t_{\mathrm{final}}$. The plots below show it is possible for $R_1\to1$ while $Q_1$ remains bounded away from zero, meaning that the tokens may synchronize to some point final point $x_\infty \not = x_0$. This does not contradict our main result, since we expect some choices of $A$ and $V$ to lead to clustering even with $k=0$. In all plots below, we show the quantities $Q_1$ and $R_1$ for various values of $k$ after allowing the system to evolve for time $t=t_{\mathrm{final}}$.

\paragraph{Setup 1: the ($A = I$, $V = J$) configuration.}

\begin{figure}[!h]
    \centering
    \includegraphics[width=\textwidth]{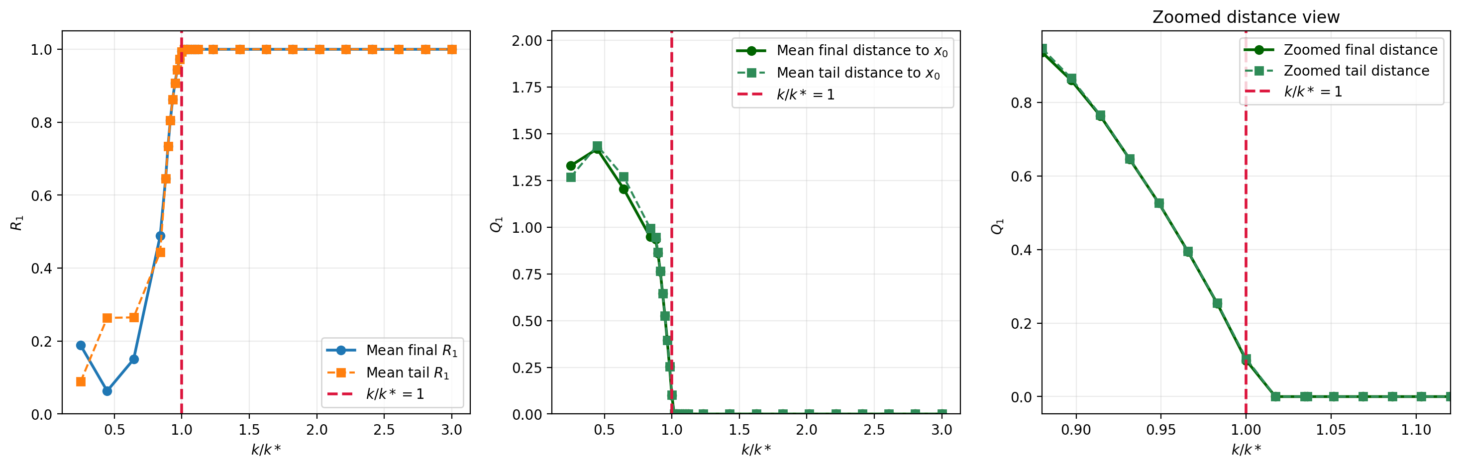}
    \caption{Setup 1 with $A=I$ and $V=J$ under softmax-normalized attention. Left: mean pairwise similarity $R_1$ as a function of $k/k^*$. Middle: mean distance to the target $x_0$. Right: zoomed in mean distance to the target $x_0$ around $k/k^*\in[0.9,1.1]$. The vertical line marks $k/k^*=1$. Experiment parameters: $d=10$, $n=100$, $t_\mathrm{final}=30$, $\beta=1$, $dt=0.001$.}
    \label{fig:sim-IJ}
\end{figure}

Figure~\ref{fig:sim-IJ} numerically verifies the transition predicted by Theorem~\ref{thm:main_result} at the critical value $k^*=1$ when we set $A = I$, $V=J$.  We are choosing $u_x(t)$ to solve~\eqref{eq:optproblem} with $k \in [0,3]$.  Here $k^* = 1$, and to stay consistent with the plots below we plot $k / k^*$.  When $k<1$, the tokens fail to cluster. As $k \to 1$, $R_1 \to 1$ and $Q_1 \to 0$, which indicates the feed-forward layer is driving the system near the target $x_0$.  Once $k > 1$, we note that $R_1 \approx 1$ and $Q_1 \approx 0$, so the system has fully synchronized about $x_0$.  The right panel of~\ref{fig:sim-IJ} suggests that the transition from asynchrony to synchrony is `sharp': if $k = k^* - \epsilon$, then numerically it appears that $Q_1 = C\epsilon$, so a small perturbation to the feed-forward layer can lead to qualitatively distinct token dynamics. Note the difference in scales for the middle and right panels.  Our theory does not apply for $k = k^*$, though we conjecture that at this critical point the system satisfies $x_i \to x_0$ on much longer timescales than for $k = k^* + \epsilon$.
\paragraph{Setup 2: random Gaussian attention.}
To test the sufficient condition in a more generic setup than above, we draw $A,V\in\mathbb{R}^{d\times d}$ with i.i.d. Gaussian entries $A_{ij},V_{ij}\sim\mathcal{N}\left(0,\frac{1}{d}\right)$. Every trial resamples $A$, $V$, $x_0$, and the initial token configuration. For each realization, we set $k^*=\|V\|_{\operatorname{op}}$ and vary the control strength relative to this value.

Unlike the experiment above, for random $A$ and $V$ Theorem~\ref{thm:main_result} does not imply synchronization must fail when $k<k^*$. The bound is sufficient i.e. for $k>k^*$ the control is strong enough to drive the tokens to $x_0$ for any $A$ and $V$. Convergence may also occur below this value for particular realizations.

\begin{figure}[!h]
    \centering
    \includegraphics[width=\textwidth]{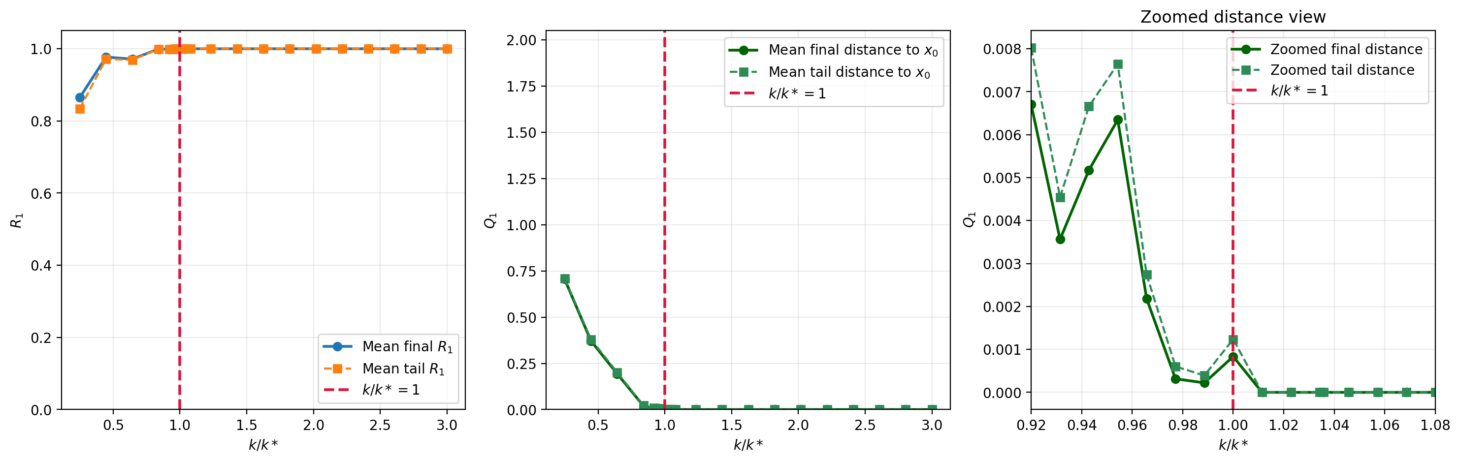}
    \caption{Setup 2, random Gaussian $A, V$. Left: mean pairwise similarity $R_1$ as a function of $k/k^*$. Middle: mean distance to the target $x_0$. Right: zoomed in mean distance to the target $x_0$ around $k/k^*\in[0.9,1.1]$. The vertical line marks $k/k^*=1$. Experiment parameters: $d=10$, $n=100$, $t_\mathrm{final}=10$, $\beta=1$, $dt=0.001$.}
    \label{fig:sim-random}
\end{figure}

Figure~\ref{fig:sim-random} shows that $R_1 \approx 1$ when $k<k^*$ which means that tokens are already synchronizing. This is expected, since clustering produced by self-attention has also been observed and studied in previous work~\citep{geshkovski2023mathematical,li2026diverse}. In our experiment $x_0$ is chosen independently from the attention matrices and the initial tokens, so there is no reason for a self-attention cluster center to be at $x_0$. As the feed-forward strength increases, the distance to $x_0$ decreases, showing that the feed-forward layer is increasingly determining the final location for token convergence. For $k>k^*$ we observe $Q_1 \approx 0$, in agreement with the sufficient condition in Theorem~\ref{thm:main_result}. As with the other experiment, we remark that our theory does not apply for $k = k^*$.  Once again, note the difference in scale for the center and right panels, and compare this to the thresholding behavior observed in the $V=J$ case above.  

\subsection{Empirical Results on Real Language Models}
\label{subsec:real_llms}

We next compare the feed-forward and attention terms in trained transformers. For each layer, we compute a feed-forward amplitude $\hat{k}$ and compare it with the theoretical quantity $k^*$. We study nanoGPT (GPT-2 small architecture, $12$ layers), Pythia-410M (GPT-NeoX architecture, $24$ layers) and TinyLlama-1.1B (Llama architecture with SwiGLU and grouped-query attention, $22$ layers).
\paragraph{Measuring $k$ for real networks.}
For a feed-forward layer $u(x)$ in the models above we wish to approximate the largest induced tangent vector, which we denote as,
\begin{equation}
    k=\max_{x\in\mathbb{S}^{d-1}}\|\operatorname{P}_x(u(x))\|.
\end{equation}
There is no closed-form expression for this maximum for a trained GELU, SiLU, or SwiGLU layer. However, the three activations above induce an infinitely differentiable field on $\S^{d-1}$.  To approximate $k$, we initialize many random unit vectors on $\S^{d-1}$ and perform gradient ascent on the sphere to identify local maxima. We report the largest value found across the random starts as $\hat{k}$.

\paragraph{Computing $k^*$ for real LLMs.} 

For a single multi-head attention layer, let $p_{ij}^{h}$ denote the softmax weights of head $h$. The contribution of the attention block to token $i$ (ignoring the projection for now) can be written as
\[
\alpha_i = 
\sum_{h=1}^{H}
W_O W_V^h
\left(
\sum_j p_{ij}^{h} x_j
\right),
\]
where $W_V^h$ corresponds to $V^h$ in Remark~\ref{rem:multihead}, and $W_O$ is a matrix that mixes the contributions of the individual heads.  Since the softmax coefficients satisfy $p_{ij}^{h}\geq 0, \sum_j p_{ij}^{h}=1,$ the vector inside each head is a convex combination of the input tokens. Thus, if $\|x_j\| < R_\ell$ \footnote{Our theory required $\|x_j\| = 1$.  This is a common idealization that facilitates analysis; real normalization schemes do not strictly enforce this constraint.},
$
\left\|
\sum_j p_{ij}^{h}x_j
\right\|
\leq R_\ell.
$
Consequently,
\[
\|\alpha_i\|
\leq
R_\ell
\sum_{h=1}^{H_\ell}
\left\|
W_O^{h}W_{V,\ell}^{h}
\right\|_{\mathrm{op}}.
\]
Motivated by the multi-head extension in Remark~\ref{rem:multihead}, we therefore define the layer-wise softmax attention bound
\begin{equation}
k_\ell^* :=
R_\ell
\sum_{h=1}^{H_\ell}
\left\|
W_{O,\ell}W_{V,\ell}^{h}
\right\|_{\mathrm{op}}.
\label{eq:kstar-real}
\end{equation}
Our plots below consider both unnormalized ($R_\ell = 1$) and normalized quantities.  Ignoring normalization gives,
\begin{equation}
k_{\ell,\mathrm{raw}}^* =
\sum_{h=1}^{H_\ell}
\left\|
W_{O,\ell}W_{V,\ell}^{h}
\right\|_{\mathrm{op}}.
\end{equation}

For Llama-style models with RMSNorm, we denote $\widetilde W_{V,\ell}^{h}$ by the matrix $W_{V, \ell}^h$ augmented with the RMSNorm operation. Then the normalized quantity used in our experiments is
\begin{equation}
k_{\ell,\mathrm{norm}}^* =
\sum_{h=1}^{H_\ell}
\left\|
W_{O,\ell}
\widetilde W_{V,\ell}
\right\|_{\mathrm{op}}.
\end{equation}

For models with LayerNorm we cannot directly augment a matrix.  Instead we estimate  $R_\ell = \sup_{\|x\|=1} \|\operatorname{LayerNorm}_\ell(x)\|$
numerically and use 
\begin{equation}k_{\ell,\mathrm{norm}}^* =
R_\ell
\sum_{h=1}^{H_\ell}
\left\|
W_{O,\ell}W_{V,\ell}^{h}
\right\|_{\mathrm{op}}.
\end{equation}

For grouped-query attention, several query heads share the same value projection. in this case we pair each query head with the value projection shared by its group and with the corresponding block of $W_O$, and its operator-norm contribution is included separately in the sum above.

\begin{figure}[h!]
  \centering  \includegraphics[width=0.85\textwidth]{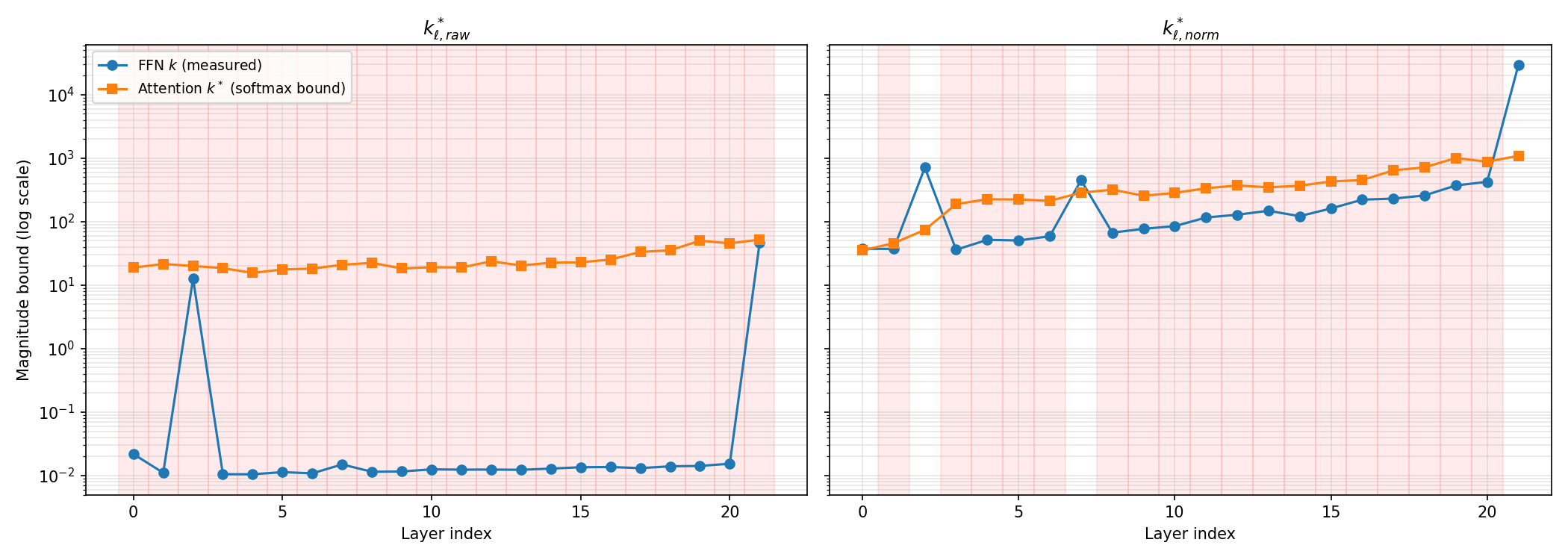}
  \caption{TinyLlama-1.1B: measured feed-forward amplitude $\hat{k}$ and softmax attention scale $k^*_{\ell,\mathrm{raw}}$ and accounting for RMSNorm $k^*_{\ell,\mathrm{norm}}$ by layer. White regions signify when the maximum FFN amplitude exceeds attention bound.}
  \label{fig:tinyllama}
\end{figure}

\paragraph{Results.}
Figures--\ref{fig:tinyllama} compare $\hat{k}$ with $k^*$ at each layer.  When $\hat{k} > k^*$, then Theorem~\ref{thm:main_result} may imply the feed-forward layer is dominating the token dynamics.  Across the three models, $\hat{k}$ and $k^*$ are often of comparable size, and in some layers we notice $\hat{k} > k^*$.  Even in regimes where Theorem~\ref{thm:main_result} does not apply, the feed-forward layers may be impacting the dynamics in a non-trivial way due.

For nanoGPT, the two quantities are in the same general range, and $\hat{k}$ is larger than $k^*$ in several early layers. In Pythia-410M, $\hat{k}$ rises above the softmax attention scale in a small group of middle layers and rises again near the end of the network. In TinyLlama, $\hat{k}$ is small through much of the network, with larger values at a few layers, including the final layer where it is close to $k^*$.

We note that $\hat{k}>k^*$ does not mean that tokens are driven to a single cluster.  Instead we interpret our results to mean there are some regions in latent space where the feed-forward layers are dominating the self-attention dynamics.  We expect the feed-forward layers are contributing to the dynamics in a very non-trivial way.  As a simple example, the multi-cluster extension in Remark~\ref{rem:multicluster} shows that the feed-forward layer can drive particles to multiple clusters, and inverting this example shows the feed-forward layers could be repelling particles in a small neighborhood.




\begin{figure}[h!]
  \centering  \includegraphics[width=0.85\textwidth]{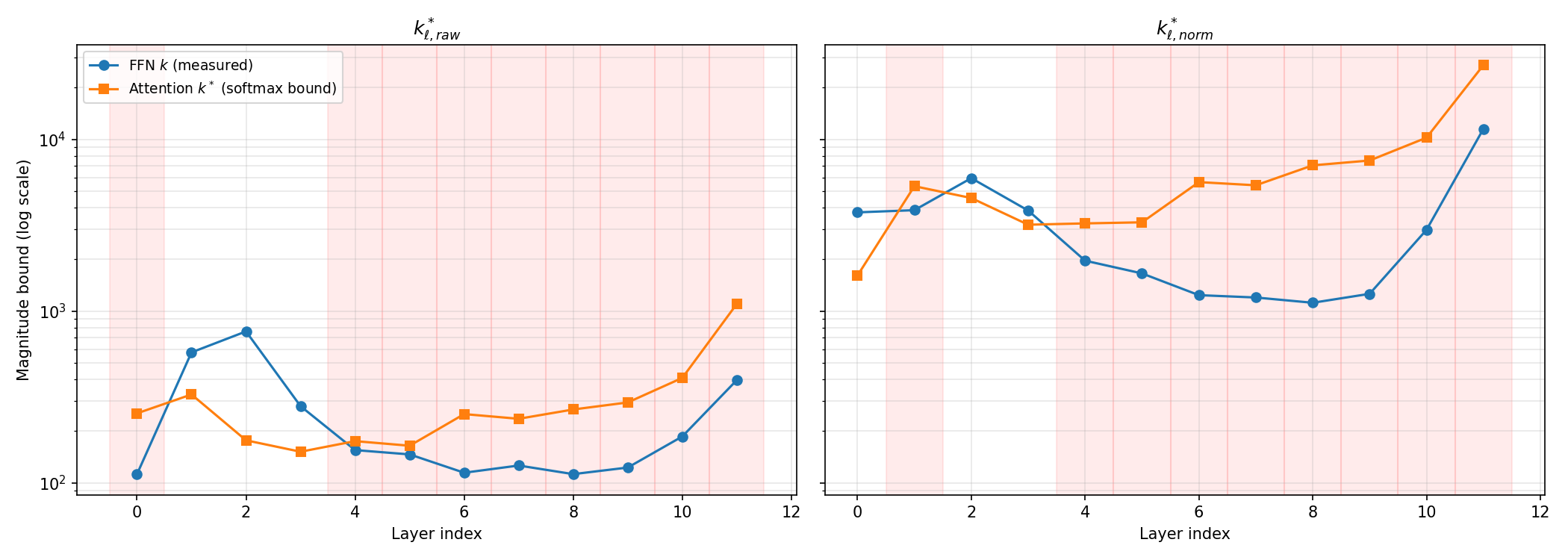}
  \caption{nanoGPT: measured feed-forward amplitude $\hat{k}$ and softmax attention scale $k^*_{\ell,\mathrm{raw}}$ and accounting for LayerNorm $k^*_{\ell,\mathrm{norm}}$ by layer. White regions signify when the maximum FFN amplitude exceeds attention bound.}
  \label{fig:nanogpt}
\end{figure}

\begin{figure}[h!]
  \centering  \includegraphics[width=0.85\textwidth]{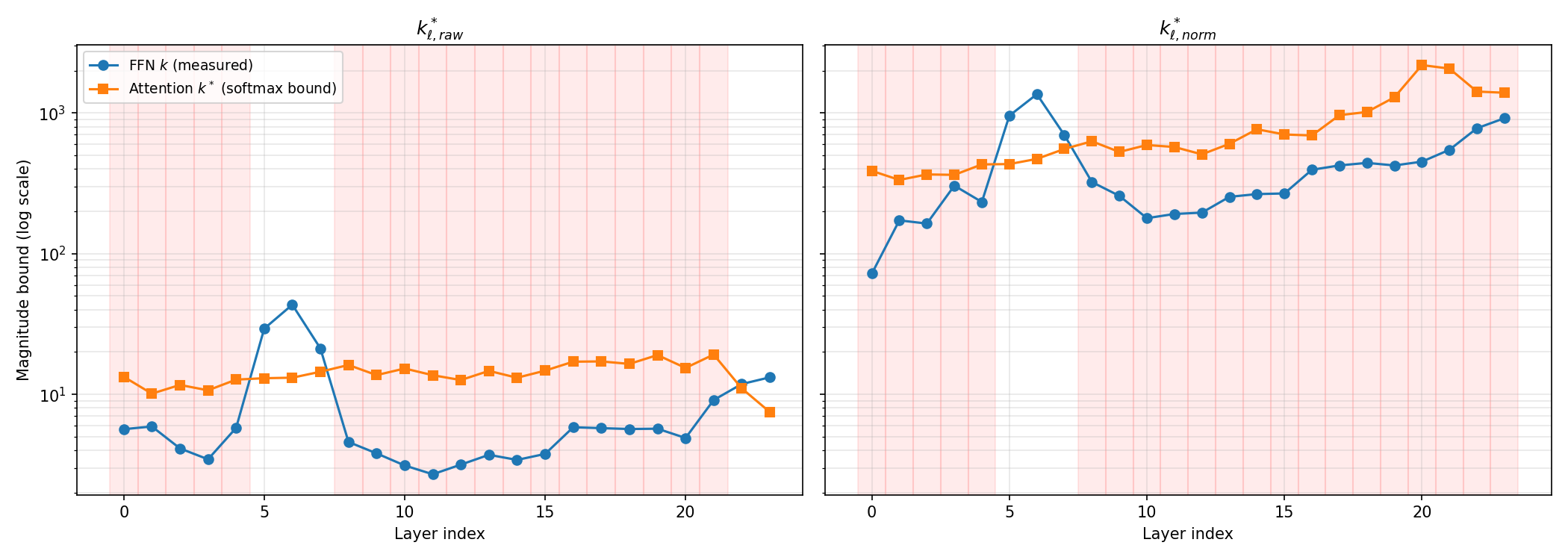}
  \caption{Pythia-410M: measured feed-forward amplitude $\hat{k}$ and softmax attention scale $k^*_{\ell,\mathrm{raw}}$ and accounting for LayerNorm $k^*_{\ell,\mathrm{norm}}$ by layer. White regions signify when the maximum FFN amplitude exceeds attention bound..}
  \label{fig:pythia}
\end{figure}

\paragraph{Limitations of the empirical study.}
Figures~\ref{fig:nanogpt}--\ref{fig:tinyllama} established that trained FFNs can attain amplitudes above $k^*$. There are a few caveats to this comparison. First, real tokens do not live exactly on $\S^{d-1}$, which is a mathematical idealization that enables our theoretical analysis. Second, these plots do not show the volume in latent space where the feed-forward layer is dominating the dynamics.  We have conducted experiments which suggest the feed-forward layer is impacting dynamics only in small regions, and this is the subject of follow-up work.  Third, even if $\hat k > k^*$, we do not expect the feed-forward layer to force tokens to a synchronized state.  We expect the influence of the feed-forward layer is complicated, and as a simple example Remark~\ref{rem:multicluster} shows clustering to multiple points is possible.  Nevertheless, our preliminary empirical study does illustrate that there is a complicated interplay between the self-attention and feed-forward layers in pre-trained LLMs.
\section{Conclusion}
We have proven that sufficiently strong feed-forward layers can independently steer tokens arbitrarily close to consensus using a control-theoretic argument.  We have also demonstrated a class of systems where `weak' feed-forward layers will always fail to force the system to synchrony.  Our numerical and empirical findings suggest that this behavior may occur in deployed LLMs.  We leave open the question of finding weak controls that can nevertheless force self-attention to synchronize about a target $x_0$.  The sufficient control problem induces $\|u_i(t)\| = k$ over nearly the whole sphere -- when can this condition be relaxed but synchrony about $x_0$ nevertheless attained?

\bibliography{biblio}
\bibliographystyle{iclr2027_conference}



\end{document}